\pdfoutput=1

\documentclass[11pt]{article}

\usepackage{acl}

\usepackage{times}
\usepackage{latexsym}

\usepackage[T1]{fontenc}

\usepackage[utf8]{inputenc}

\usepackage{microtype}
\usepackage{amsmath, amssymb}
\usepackage{booktabs}
\usepackage{graphicx}
\usepackage{multirow}

\title{Incorporating Hierarchy into Text Encoder: a Contrastive Learning Approach for Hierarchical Text Classification}

\author{
Zihan Wang, Peiyi Wang, Lianzhe Huang, Xin Sun, Houfeng Wang$\footnotemark[1]$
\\ 
Key Laboratory of Computational Linguistics, Peking University, MOE, China \\
 \texttt{\{wangzh9969, wangpeiyi9979\}@gmail.com} \\
 \texttt{\{hlz, sunx5, wanghf\}@pku.edu.cn}
}

\begin{document}
\maketitle
\renewcommand{\thefootnote}{\fnsymbol{footnote}}
\footnotetext[1]{Corresponding author.}
\renewcommand{\thefootnote}{\arabic{footnote}}

\begin{abstract}
Hierarchical text classification is a challenging subtask of multi-label classification due to its complex label hierarchy. Existing methods encode text and label hierarchy separately and mix their representations for classification, where the hierarchy remains unchanged for all input text. Instead of modeling them separately, in this work, we propose Hierarchy-guided Contrastive Learning (HGCLR) to directly embed the hierarchy into a text encoder. During training, HGCLR constructs positive samples for input text under the guidance of the label hierarchy. By pulling together the input text and its positive sample, the text encoder can learn to generate the hierarchy-aware text representation independently. Therefore, after training, the HGCLR enhanced text encoder can dispense with the redundant hierarchy. Extensive experiments on three benchmark datasets verify the effectiveness of HGCLR.


\end{abstract}

\section{Introduction}
Hierarchical Text Classification (HTC) aims to categorize text into a set of labels that are organized in a structured hierarchy \cite{silla2011survey}. The taxonomic hierarchy is commonly modeled as a tree or a directed acyclic graph, in which each node is a label to be classified. As a subtask of multi-label classification, the key challenge of HTC is how to model the large-scale, imbalanced, and structured label hierarchy \cite{mao2019hierarchical}.

\begin{figure}[t]
\centering
\includegraphics[width=\linewidth]{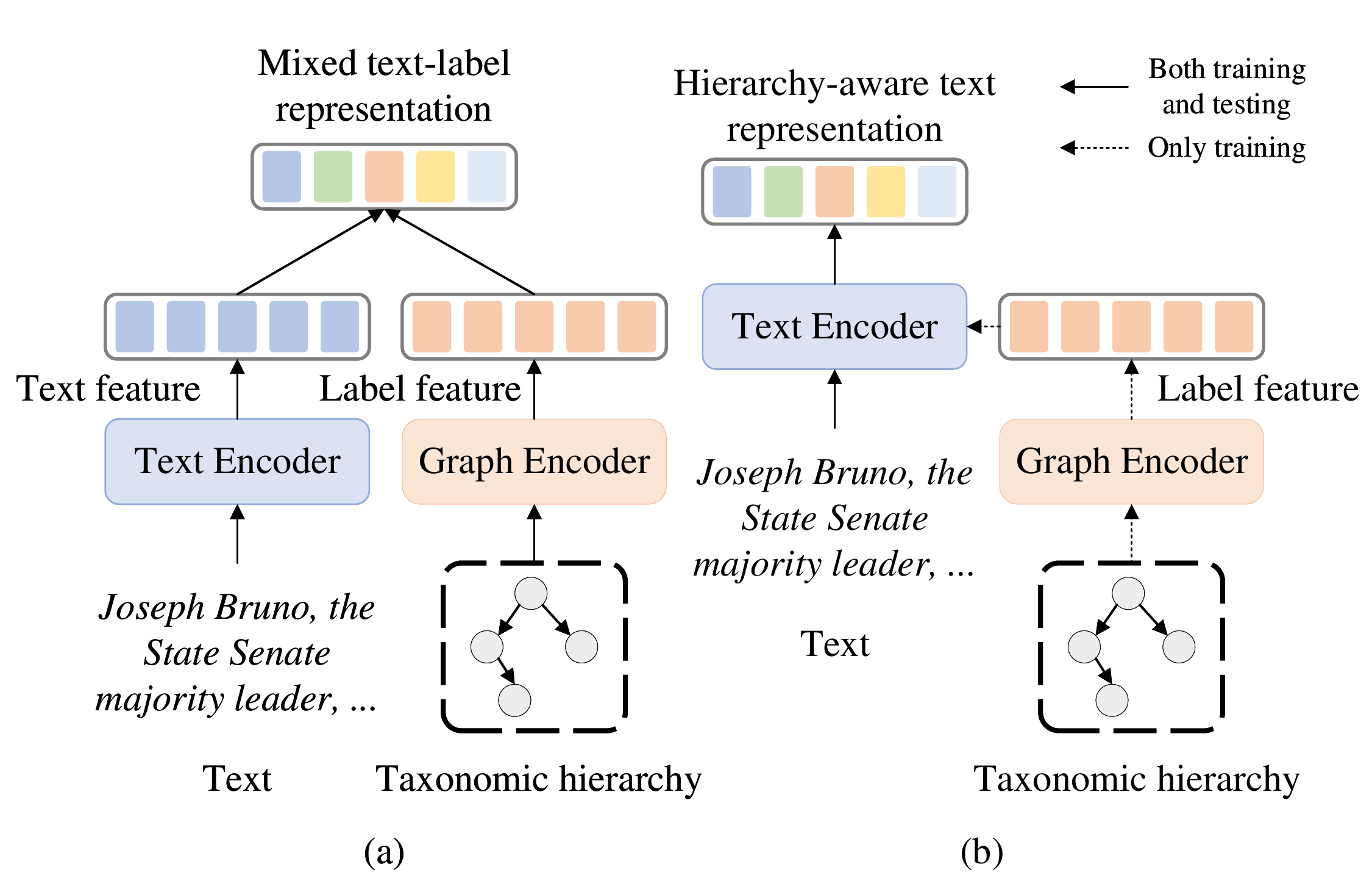}
\caption{Two ways of introducing hierarchy information. (a) Previous work model text and labels separately and find a mixed representation. (b) Our method incorporating hierarchy information into text encoder for a hierarchy-aware text representation. }
\label{fig:1}
\end{figure}

The existing methods of HTC have variously introduced hierarchical information. Among recent researches, the state-of-the-art models encode text and label hierarchy separately and aggregate two representations before being classified by a mixed feature \cite{zhou2020hierarchy, deng2021htcinfomax}. As denoted in the left part of Figure \ref{fig:1}, their main goal is to sufficiently interact between text and structure to achieve a mixed representation \cite{chenhierarchy}, which is highly useful for classification \cite{chen2020hyperbolic}. However, since the label hierarchy remains unchanged for all text inputs, the graph encoder provides exactly the same representation regardless of the input. Therefore, the text representation interacts with constant hierarchy representation and thus the interaction seems redundant and less effective. Alternatively, we attempt to inject the constant hierarchy representation into the text encoder. So that after being fully trained, a hierarchy-aware text representation can be acquired without the constant label feature. As in the right part of Figure \ref{fig:1}, instead of modeling text and labels separately, migrating label hierarchy into text encoding may benefit HTC by a proper representation learning method.

To this end, we adopt contrastive learning for the hierarchy-aware representation. Contrastive learning, which aims to concentrate positive samples and push apart negative samples, has been considered as effective in constructing meaningful representations \cite{kim2021self}. Previous work on contrastive learning illustrates that it is critical to building challenging samples \cite{alzantot2018generating, wang2021cline, tan2020s, wu2020clear}. For multi-label classification, we attempt to construct high-quality positive examples. Existing methods for positive example generation includes data augmentation \cite{meng2021coco, wu2020clear}, dropout \cite{gao2021simcse}, and adversarial attack \cite{wang2021cline, pan2021improved}. These techniques are either unsupervised or task-unspecific: the generation of positive samples has no relation with the HTC task and thus are incompetent to acquire hierarchy-aware representations. As mentioned, we argue that both the ground-truth label as well as the taxonomic hierarchy should be considered for the HTC task.

To construct positive samples which are both label-guided and hierarchy-involved, our approach is motivated by a preliminary observation. Notice that when we classify text into a certain category, most words or tokens are not important. For instance, when a paragraph of news report about a lately sports match is classified as ``basketball'', few keywords like ``NBA'' or ``backboard'' have large impacts while the game result has less influence. So, given a sequence and its labels, a shorten sequence that only keeps few keywords should maintain the labels. In fact, this idea is similar to adversarial attack, which aims to find ``important tokens'' which affect classification most \cite{zhang2020adversarial}. The difference is that adversarial attack tries to modify ``important tokens'' to fool the model, whereas our approach modifies ``unimportant tokens'' to keep the classification result unchanged.

Under such observation, we construct positive samples as pairs of input sequences and theirs shorten counterparts, and propose Hierarchy-Guided Contrastive Learning (HGCLR) for HTC. In order to locate keywords under given labels, we directly calculate the attention weight of each token embedding on each label, and tokens with weight above a threshold are considered important to according label. We use a graph encoder to encode label hierarchy and output label features. Unlike previous studies with GCN or GAT, we modify a Graphormer \cite{ying2021transformers} as our graph encoder. Graphormer encodes graphs by Transformer blocks and outperforms other graph encoders on several graph-related tasks. It models the graph from multiple dimensions, which can be customized easily for HTC task.

The main contribution of our work can be summarized as follows:
\begin{itemize}
    \item We propose Hierarchy-Guided Contrastive Learning (HGCLR) to obtain hierarchy-aware text representation for HTC. To our knowledge, this is the first work that adopts contrastive learning on HTC. 
    \item For contrastive learning, we construct positive samples by a novel approach guided by label hierarchy. The model employs a modified Graphormer, which is a new state-of-the-art graph encoder.
    \item Experiments demonstrate that the proposed model achieves improvements on three datasets. Our code is available at \href{https://github.com/wzh9969/contrastive-htc}{https://github.com/wzh9969/contrastive-htc}.
\end{itemize}

\section{Related Work}
\subsection{Hierarchical Text Classification}
Existing work for HTC could be categorized into local and global approaches based on their ways of treating the label hierarchy \cite{zhou2020hierarchy}. Local approaches build classifiers for each node or level while the global ones build only one classifier for the entire graph. \citet{banerjee2019hierarchical} builds one classifier per label and transfers parameters of the parent model for child models. \citet{wehrmann2018hierarchical} proposes a hybrid model combining local and global optimizations. \citet{shimura2018hft} applies CNN to utilize the data in the upper levels to contribute categorization in the lower levels.

The early global approaches neglect the hierarchical structure of labels and view the problem as a flat multi-label classification \cite{johnson2014effective}. Later on, some work tries to coalesce the label structure by recursive regularization \cite{gopal2013recursive}, reinforcement learning \cite{mao2019hierarchical}, capsule network \cite{peng2019hierarchical}, and meta-learning \cite{wu2019learning}. Although such methods can capture the hierarchical information, recent researches demonstrate that encoding the holistic label structure directly by a structure encoder can further improve performance. \citet{zhou2020hierarchy} designs a structure encoder that integrates the label prior hierarchy knowledge to learn label representations. \citet{chen2020hyperbolic} embeds word and label hierarchies jointly in the hyperbolic space. \citet{zhang2020hcn} extracts text features according to different hierarchy levels. \citet{deng2021htcinfomax} introduces information maximization to constrain label representation learning. \citet{zhao2021hierarchical} designs a self-adaption fusion strategy to extract features from text and label. \citet{chenhierarchy} views the problem as semantic matching and tries BERT as text encoder. \citet{wang2021cognitive} proposes a cognitive structure learning model for HTC. Similar to other work, they model text and label separately.


\subsection{Contrastive Learning}
Contrastive learning is originally proposed in Computer Vision (CV) as a weak-supervised representation learning method. Works such as MoCo \cite{he2020momentum} and SimCLR \cite{chen2020simple} have bridged the gap between self-supervised learning and supervised learning on multiple CV datasets. A key component for applying contrastive learning on NLP is how to build positive pairs \cite{pan2021improved}. Data augmentation techniques such as back-translation \cite{fang2020cert}, word or span permutation \cite{wu2020clear}, and random masking \cite{meng2021coco} can generate pair of data with similar meanings. \citet{gao2021simcse} uses different dropout masks on the same data to generate positive pairs. \citet{kim2021self} utilizes BERT representation by a fixed copy of BERT. These methods do not rely on downstream tasks while some researchers leverage supervised information for better performance on text classification. \citet{wang2021cline} constructs both positive and negative pairs especially for sentimental classification by word replacement. \citet{pan2021improved} proposes to regularize Transformer-based encoders for text classification tasks by FGSM \cite{goodfellow2014explaining}, an adversarial attack method based on gradient. Though methods above are designed for classification, the construction of positive samples hardly relies on their categories, neglecting the connection and diversity between different labels. For HTC, the taxonomic hierarchy models the relation between labels, which we believe can help positive sample generation.

\section{Problem Definition}
Given a input text $x = \{x_1,x_2,...,x_n\}$, Hierarchical Text Classification (HTC) aims to predict a subset $y$ of label set $Y$, where $n$ is the length of the input sequence and $k$ is the size of set $Y$. The candidate labels $y_i\in Y$ are predefined and organized as a Directed Acyclic Graph (DAG) $G=(Y,E)$, where node set $Y$ are labels and edge set $E$ denotes their hierarchy. For simplicity, we do not distinguish a label with its node in the hierarchy so that $y_i$ is both a label and a node. Since a non-root label of HTC has one and only one father, the taxonomic hierarchy can be converted to a tree-like hierarchy. The subset $y$ corresponds to one or more paths in $G$: for any non-root label $y_j\in y$, a father node (label) of $y_j$ is in the subset $y$.

\section{Methodology}
\begin{figure*}[t]
\centering
\includegraphics[width=\linewidth]{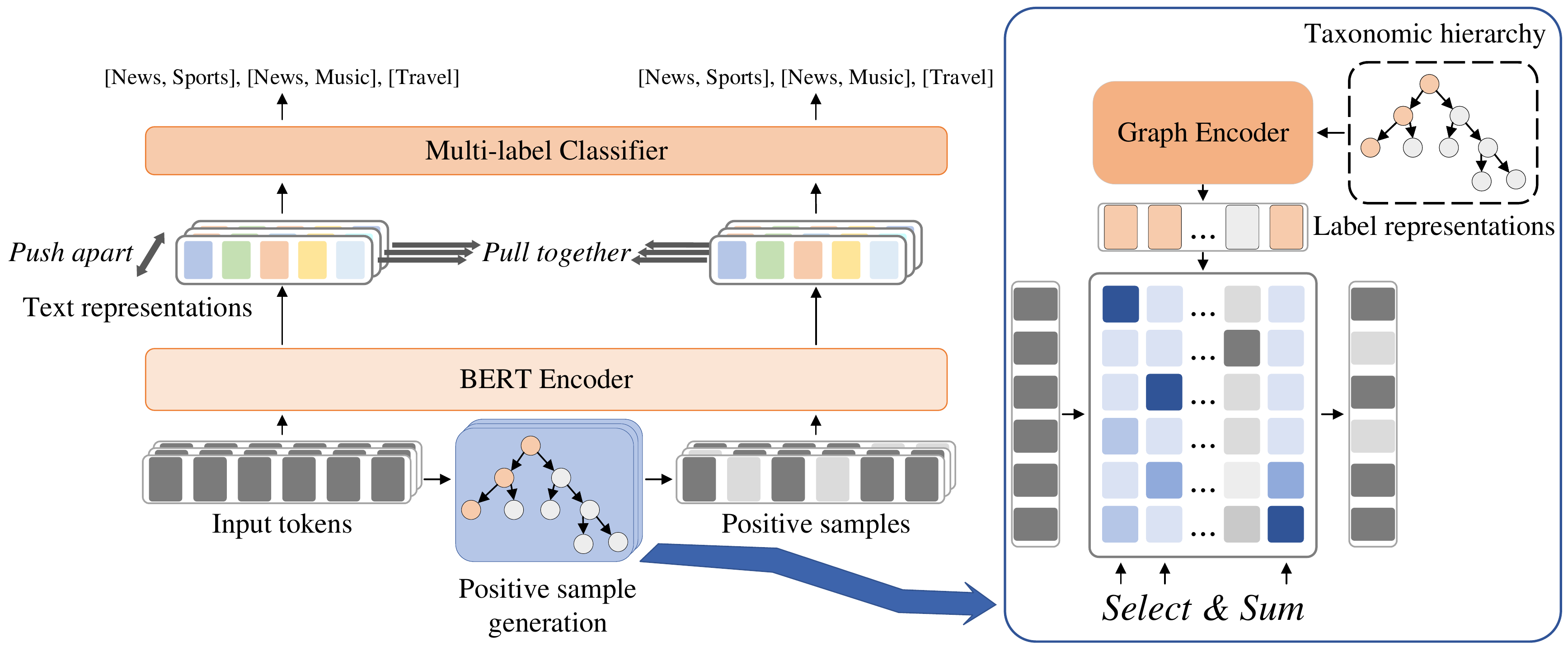}
\caption{An overview of HGCLR under a batch of 3. HGCLR adopts a contrastive learning framework to regularize BERT representations. We construct positive samples by masking unimportant tokens under the guidance of hierarchy and labels. By pulling together and pushing apart representations, the hierarchy information can be injected into the BERT encoder.}
\label{fig:2}
\end{figure*}

In this section, we will describe the proposed HGCLR in detail. Figure \ref{fig:2} shows the overall architecture of the model.

\subsection{Text Encoder}
Our approach needs a strong text encoder for hierarchy injection, so we choose BERT \cite{devlin2018bert} as the text encoder.
Given an input token sequence:
\begin{equation} \label{eq:1}
    x = \{{\rm [CLS]}, x_1,x_2,...,x_{n-2}, {\rm [SEP]}\}
\end{equation}
where $\rm [CLS]$ and $\rm [SEP]$ are two special tokens indicating the beginning and the end of the sequence, the input is fed into BERT. For convenience, we denote the length of the sequence as $n$. The text encoder outputs hidden representation for each token:
\begin{equation}
    H=\rm{BERT}(x)
\end{equation}
where $H\in \mathbb{R}^{n\times d_h}$ and $d_h$ is the hidden size. We use the hidden state of the first token ($\rm [CLS]$) for representing the whole sequence $h_x=h_{\rm [CLS]}$.

\subsection{Graph Encoder} \label{ch:1}
We model the label hierarchy with a customized Graphormer \cite{ying2021transformers}. Graphormer models graphs on the base of Transformer layer \cite{vaswani2017attention} with spatial encoding and edge encoding, so it can leverage the most powerful sequential modeling network in the graph domain. We organize the original feature for node $y_i$ as the sum of label embedding and its name embedding:
\begin{equation}
    f_i={\rm label\_emb}(y_i)+{\rm name\_emb}(y_i).
\end{equation}
Label embedding is a learnable embedding that takes a label as input and outputs a vector with size $d_h$. Name embedding takes the advantage of the name of the label, which we believe contains fruitful information as a summary of the entire class. We use the average of BERT token embedding of the label as its name embedding, which also has a size of $d_h$. Unlike previous work which only adopts names on initialization, we share embedding weights across text and labels to make label features more instructive. With all node features stack as a matrix $F\in \mathbb{R}^{k\times d_h}$, a standard self-attention layer can then be used for feature migration.

To leverage the structural information, spatial encoding and edge encoding modify the Query-Key product matrix $A^G$ in the self-attention layer:
\begin{equation} \label{eq:5}
    A^G_{ij} = \frac{(f_iW_Q^G)(f_jW_K^G)^T}{\sqrt{d_h}}+c_{ij} + b_{\phi(y_i,y_j)}
\end{equation}
where $c_{ij}=\frac1D\sum_{n=1}^Dw_{e_n}$ and $D=\phi(y_i,y_j)$.
The first term in Equation \ref{eq:5} is the standard scale-dot attention, and query and key are projected by $W_Q^G\in\mathbb{R}^{d_h\times d_h}$ and $W_K^G\in\mathbb{R}^{d_h\times d_h}$. $c_{ij}$ is the edge encoding and $\phi(y_i,y_j)$ denotes the distance between two nodes $y_i$ and $y_j$. Since the graph is a tree in our problem, for node $y_i$ and $y_j$, one and only one path $(e_1, e_2,...,e_D)$ can be found between them in the underlying graph $G'$ so that $c_{ij}$ denotes the edge information between two nodes and $w_{e_i}\in\mathbb{R}^1$ is a learnable weight for each edge. $b_{\phi(y_i,y_j)}$ is the spatial encoding, which measures the connectivity between two nodes. It is a learnable scalar indexed by $\phi(y_i,y_j)$.

The graph-involved attention weight matrix $A^G$ is then followed by Softmax, multiplying with value matrix and residual connection \& layer normalization to calculate the self-attention,
\begin{equation}
    L=\rm LayerNorm({\rm softmax}(A^G)V+F)
\end{equation}
We use $L$ as the label feature for the next step. The Graphormer we use is a variant of the self-attention layer, for more details on the full structure of Graphormer, please refer to the original paper.

\subsection{Positive Sample Generation} \label{ch:2}
As mentioned, the goal for the positive sample generation is to keep a fraction of tokens while retaining the labels. Given a token sequence as Equation \ref{eq:1}, the token embedding of BERT is defined as:
\begin{equation} \label{eq:6}
    \{e_1,e_2,...,e_n\}={\rm BERT\_emb}(x)
\end{equation}

The scale-dot attention weight between token embedding and label feature is first calculated to determine the importance of a token on a label,
\begin{equation} \label{eq:7}
    q_i=e_iW_Q,k_j=l_jW_K,A_{ij}=\frac{q_ik_j^T}{\sqrt{d_h}}
\end{equation}
The query and key are token embeddings and label features respectively, and $W_Q\in \mathbb{R}^{d_h\times d_h}$ and $W_K\in \mathbb{R}^{d_h\times d_h}$ are two weight matrices. Thus, for a certain $x_i$, its probability of belonging to label $y_j$ can be normalized by a Softmax function.

Next, given a label $y_j$, we can sample key tokens from that distribution and form a positive sample $\hat{x}$. To make the sampling differentiable, we replace the Softmax function with Gumbel-Softmax \cite{jang2016categorical} to simulate the sampling operation:
\begin{equation}
    P_{ij}={\rm gumbel\_softmax}(A_{i1}, A_{i2}, ..., A_{ik})_j
\end{equation}
Notice that a token can impact more than one label, so we do not discretize the probability as one-hot vectors in this step. Instead, we keep tokens for positive examples if their probabilities of being sampled exceed a certain threshold $\gamma$, which can also control the fraction of tokens to be retrained. For multi-label classification, we simply add the probabilities of all ground-truth labels and obtain the probability of a token $x_i$ regarding its ground-truth label set $y$ as:
\begin{equation}
    P_i=\sum_{j\in y}P_{ij}
\end{equation}

Finally, the positive sample $\hat{x}$ is constructed as:
\begin{equation}\label{eq:2}
    \hat{x}=\{x_i \text{ if } P_i>\gamma \text{ else } \mathbf{0} \}
\end{equation}
where $\mathbf{0}$ is a special token that has an embedding of all zeros so that key tokens can keep their positions. The select operation is not differentiable, so we implement it differently to make sure the whole model can be trained end-to-end. Details are illustrated in Appendix \ref{app:1}.

The positive sample is fed to the same BERT as the original one,
\begin{equation}
    \hat{H}={\rm BERT}(\hat{x})
\end{equation}
and get a sequence representation $\hat{h_{x}}$ with the first token before being classified. We assume the positive sample should retain the labels, so we use classification loss of the positive sample as a guidance of the graph encoder and the positive sample generation.

\subsection{Contrastive Learning Module}
Intuitively, given a pair of token sequences and their positive counterpart, their encoded sentence-level representation should be as similar to each other as possible. Meanwhile, examples not from the same pair should be farther away in the representation space. 

Concretely, with a batch of $N$ hidden state of positive pairs $(h_i, \hat{h_i})$, we add a non-linear layer on top of them:
\begin{equation}
\begin{aligned}
   &c_i=W_2{\rm ReLU}(W_1h_i) \\
   &\hat{c_i}=W_2{\rm ReLU}(W_1\hat{h_i})
\end{aligned}
\end{equation}
where $W_1\in\mathrm{R}^{d_h\times d_h}$, $W_2\in\mathrm{R}^{d_h\times d_h}$. For each example, there are $2(N-1)$ negative pairs, i.e., all the remaining examples in the batch are negative examples. Thus, for a batch of $2N$ examples $\mathbf{Z}=\{z\in\{c_i\}\cup \{\hat{c_i}\}\}$, we compute the NT-Xent loss \cite{chen2020simple} for $z_m$ as:
\begin{equation}
    L_m^{con}=-\log\frac{\exp ({\rm sim}(z_m, \mu(z_m))/\tau)}{\sum^{2N}_{i=1,i\not= m}\exp ({\rm sim}(z_m, z_i)/\tau)}
\end{equation}
where $\rm sim$ is the cosine similarity function as ${\rm sim}(u, v)=u\cdot v/ \|u\|\|v\|$ and $\mu$ is a matching function as:
\begin{equation}
    \mu(z_m)=\left\{ \begin{array}{ll}
         c_i, \text{ if } z_m=\hat{c_i}&\\
         \hat{c_i}, \text{ if } z_m=c_i&
    \end{array}
    \right.
\end{equation}
$\tau$ is a temperature hyperparameter.

The total contrastive loss is the mean loss of all examples:
\begin{equation}
    L^{con}=\frac{1}{2N}\sum_{m=1}^{2N}L_m^{con}
\end{equation}

\subsection{Classification and Objective Function}
Following previous work \cite{zhou2020hierarchy}, we flatten the hierarchy for multi-label classification. The hidden feature is fed into a linear layer, and a sigmoid function is used for calculating the probability. The probability of text $i$ on label $j$ is:
\begin{equation} \label{eq:3}
    p_{ij}={\rm sigmoid}(W_{c}h_i+b_c)_j
\end{equation}
where $W_C\in\mathbb{R}^{k\times d_h}$ and $b_c \in \mathbb{R}^{k}$ are weights and bias.
For multi-label classification, we use a binary cross-entropy loss function for text $i$ on label $j$,
\begin{equation}
L_{ij}^C=-y_{ij}\log(p_{ij})-(1-y_{ij})\log(1-p_{ij})
\end{equation}
\begin{equation} \label{eq:4}
    L^C=\sum_{i=1}^N\sum_{j=1}^kL_{ij}^C
\end{equation}
where $y_{ij}$ is the ground truth. The classification loss of the constructed positive examples $\hat{L^C}$ can be calculated similarly by Equation \ref{eq:3} and Equation \ref{eq:4} with $\hat{h_i}$ substituting for $h_i$.

The final loss function is the combination of classification loss of original data, classification loss of the constructed positive samples, and the contrastive learning loss:
\begin{equation}
    L=L^C+\hat{L^C}+\lambda L^{con}
\end{equation}
where $\lambda$ is a hyperparameter controlling the weight of contrastive loss.

During testing, we only use the text encoder for classification and the model degenerates to a BERT encoder with a classification head.

\begin{table}[t]
\resizebox{\linewidth}{!}{
\begin{tabular}{c|cccccc}
\toprule
Dataset & $|Y|$   & Depth &   Avg($|y_i|$)   & Train  & Dev   & Test    \\ \midrule
WOS     & 141 & 2     & 2.0  & 30,070 & 7,518 & 9,397   \\
NYT     & 166 & 8     & 7.6  & 23,345 & 5,834 & 7,292   \\
RCV1-V2    & 103 & 4     & 3.24 & 20,833 & 2,316 & 781,265 \\ \bottomrule
\end{tabular}
}
\caption{Data Statistics. $|Y|$ is the number of classes. Depth is the maximum level of hierarchy. Avg($|y_i|$) is the average number of classes per sample.}
\label{tab:1}
\end{table}

\begin{table*}[t]
\resizebox{\textwidth}{!}{%
\begin{tabular}{@{}ccccccc@{}}
\toprule
\multirow{2}{*}{Model}          & \multicolumn{2}{c}{WOS} & \multicolumn{2}{c}{NYT} & \multicolumn{2}{c}{RCV1-V2} \\ \cmidrule(l){2-7} 
                                & Micro-F1   & Macro-F1   & Micro-F1   & Macro-F1   & Micro-F1     & Macro-F1     \\ \midrule
\multicolumn{7}{c}{\textbf{Hierarchy-Aware Models}}                                                                        \\ \midrule
TextRCNN \cite{zhou2020hierarchy}& 83.55      & 76.99      & 70.83      & 56.18      & 81.57        & 59.25        \\
HiAGM \cite{zhou2020hierarchy}& 85.82      & 80.28      & 74.97      & 60.83      & 83.96        & 63.35        \\
HTCInfoMax \cite{deng2021htcinfomax}& 85.58      & 80.05      & -          & -          & 83.51        & 62.71        \\
HiMatch \cite{chenhierarchy}   & 86.20      & 80.53      & -          & -          & 84.73        & 64.11        \\ \midrule
\multicolumn{7}{c}{\textbf{Pretrained Language Models}}                                                                    \\ \midrule
BERT (Our implement)            & 85.63      & 79.07      & 78.24      & 65.62      & 85.65        & 67.02        \\
BERT \cite{chenhierarchy}     & 86.26      & 80.58      & -          & -          & 86.26        & 67.35        \\
BERT+HiAGM (Our implement)      & 86.04      &  80.19     & 78.64     & 66.76      &  85.58      & 67.93        \\
BERT+HTCInfoMax (Our implement) &  86.30     &  79.97     & 78.75          & 67.31          &  85.53      &  67.09      \\
BERT+HiMatch \cite{chenhierarchy}& 86.70      & 81.06      & -          & -          & 86.33        & \textbf{68.66}        \\
HGCLR                     &  \textbf{87.11}    &   \textbf{81.20}   &   \textbf{78.86}   &   \textbf{67.96}  &   \textbf{86.49}      &  68.31      \\ \bottomrule
\end{tabular}%
}
\caption{Experimental results of our proposed model on several datasets. For a fair comparison, we implement some baseline with BERT encoder. We cannot reproduce the BERT results reported in \citet{chenhierarchy} so that we also report the results of our version of BERT.}
\label{tab:2}
\end{table*}

\section{Experiments}
\subsection{Experiment Setup}

\paragraph{Datasets and Evaluation Metrics}
We experiment on Web-of-Science (WOS) \cite{kowsari2017hdltex}, NYTimes (NYT) \cite{sandhaus2008new}, and RCV1-V2 \cite{lewis2004rcv1} datasets for comparison and analysis. WOS contains abstracts of published papers from Web of Science while NYT and RCV1-V2 are both news categorization corpora. We follow the data processing of previous work \cite{zhou2020hierarchy}. WOS is for single-path HTC while NYT and RCV1-V2 include multi-path taxonomic labels. The statistic details are illustrated in Table \ref{tab:1}.
Similar to previous work, We measure the experimental results with Macro-F1 and Micro-F1. 

\paragraph{Implement Details}  For text encoder, we use \texttt{bert-base-uncased} from Transformers \cite{wolf-etal-2020-transformers} as the base architecture. Notice that we denote the attention layer in Eq. \ref{eq:5} and Eq. \ref{eq:7} as single-head attentions but they can be extended to multi-head attentions as the original Transformer block. For Graphormer, we set the attention head to $8$ and feature size $d_h$ to $768$. The batch size is set to $12$. The optimizer is Adam with a learning rate of $3e-5$. We implement our model in PyTorch and train end-to-end. We train the model with train set and evaluate on development set after every epoch, and stop training if the Macro-F1 does not increase for $6$ epochs. The threshold $\gamma$ is set to $0.02$ on WOS and $0.005$ on NYT and RCV1-V2. The loss weight $\lambda$ is set to $0.1$ on WOS and RCV1-V2 and $0.3$ on NYT. $\gamma$ and $\lambda$ are selected by grid search on development set. The temperature of contrastive module is fixed to $1$ since we have achieved promising results with this default setting in preliminary experiments.

\paragraph{Baselines}
We select a few recent work as baselines. HiAGM \cite{zhou2020hierarchy}, HTCInfoMax \cite{deng2021htcinfomax}, and HiMatch \cite{chenhierarchy} are a branch of work that propose fusion strategies for mixed text-hierarchy representation. HiAGM applies soft attention over text feature and label feature for the mixed feature. HTCInfoMax improves HiAGM by regularizing the label representation with a prior distribution. HiMatch matches text representation with label representation in a joint embedding space and uses joint representation for classification. HiMatch is the state-of-the-art before our work. All approaches except HiMatch adopt TextRCNN \cite{lai2015recurrent} as text encoder so that we implement them with BERT for a fair comparison.

\subsection{Experimental Results}
Main results are shown in Table \ref{tab:2}. Instead of modeling text and labels separately, our model can make more use of the strong text encoder by migrating hierarchy information directly into BERT encoder. On WOS, the proposed HGCLR can achieve 1.5\% and 2.1\% improvement on Micro-F1 and Macro-F1 respectively comparing to BERT and is better than HiMatch even if its base model has far better performance.

BERT was trained on news corpus so that the base model already has decent performance on NYT and RCV1-V2, outperforming post-pretrain models by a large amount. On NYT, our approach observes a 2.3\% boost on Macro-F1 comparing to BERT while sightly increases on Micro-F1 and outperform previous methods on both measurements.

On RCV1-V2, all baselines hardly improve Micro-F1 and only influence Macro-F1 comparing to BERT. HTCInfoMax experiences a decrease because its constraint on text representation may contradict with BERT on this dataset. HiMatch behaves extremely well on RCV1-V2 with Macro-F1 as measurement while our approach achieves state-of-the-art on Micro-F1. Besides the potential implement difference on BERT encoder, RCV1-V2 dataset provides no label name, which invalids our name embedding for label representation. Baselines like HiAGM and HiMatch only initialize labl embedding with their names so that this flaw has less impact. We will discuss more on name embedding in next section.
\subsection{Analysis}

\begin{table}[t]
\centering
\resizebox{0.9\linewidth}{!}{
\begin{tabular}{lcc}
\toprule
Ablation Models   & Micro-F1 & Macro-F1 \\ \midrule
BERT              & 85.75    & 79.36    \\ \midrule
HGCLR             & \textbf{87.46}    & \textbf{81.52}    \\ 
-\textit{r.p.} GCN           &   87.06  &   80.63  \\
-\textit{r.p.} GAT           &   87.18  &   81.45  \\
-\textit{r.m.} graph encoder    &   86.67  &   80.11  \\
-\textit{r.m.} contrastive loss & 86.72    & 80.97    \\ \bottomrule
\end{tabular}}
\caption{Performance when replace or remove some components of HGCLR on the development set of WOS. \textit{r.p.} stands for \textit{replace} and \textit{r.m.} stands for \textit{remove}. We remove the contrastive loss by setting $\lambda=0$.}
\label{tab:3}

\end{table}
The main differences between our work and previous ones are the graph encoder and contrastive learning. To illustrate the effectiveness of these two parts, we test our model with them replaced or removed. We report the results on the development set of WOS for illustration. We first replace Graphormer with GCN and GAT (\textit{r.p.} GCN and \textit{r.p.} GAT), results are in Table \ref{tab:3}. We find that Graphormer outperforms both graph encoders on this task. GAT also involves the attention mechanism but a node can only attend to its neighbors. Graphormer adopts global attention where each node can attend to all others in the graph, which is proven empirically more effective on this task. When the graph encoder is removed entirely (-\textit{r.m.} graph encoder), the results drop significantly, showing the necessity of incorporating graph encoder for HTC task.

The model without contrastive loss is similar to a pure data augmentation approach, where positive examples stand as augment data. As the last row of Table \ref{tab:3}, on development set, both the positive pair generation strategy and the contrastive learning framework have contributions to the model. Our data generation strategy is effective even without contrastive learning, improving BERT encoder by around 1\% on two measurements. Contrastive learning can further boost performance by regularizing text representation.

We further analyze the effect of incorporating label hierarchy, the Graphormer, and the positive samples generation strategy in detail.

\subsubsection{Effect of Hierarchy}

\begin{figure}[t]
\centering
\includegraphics[width=0.95\linewidth]{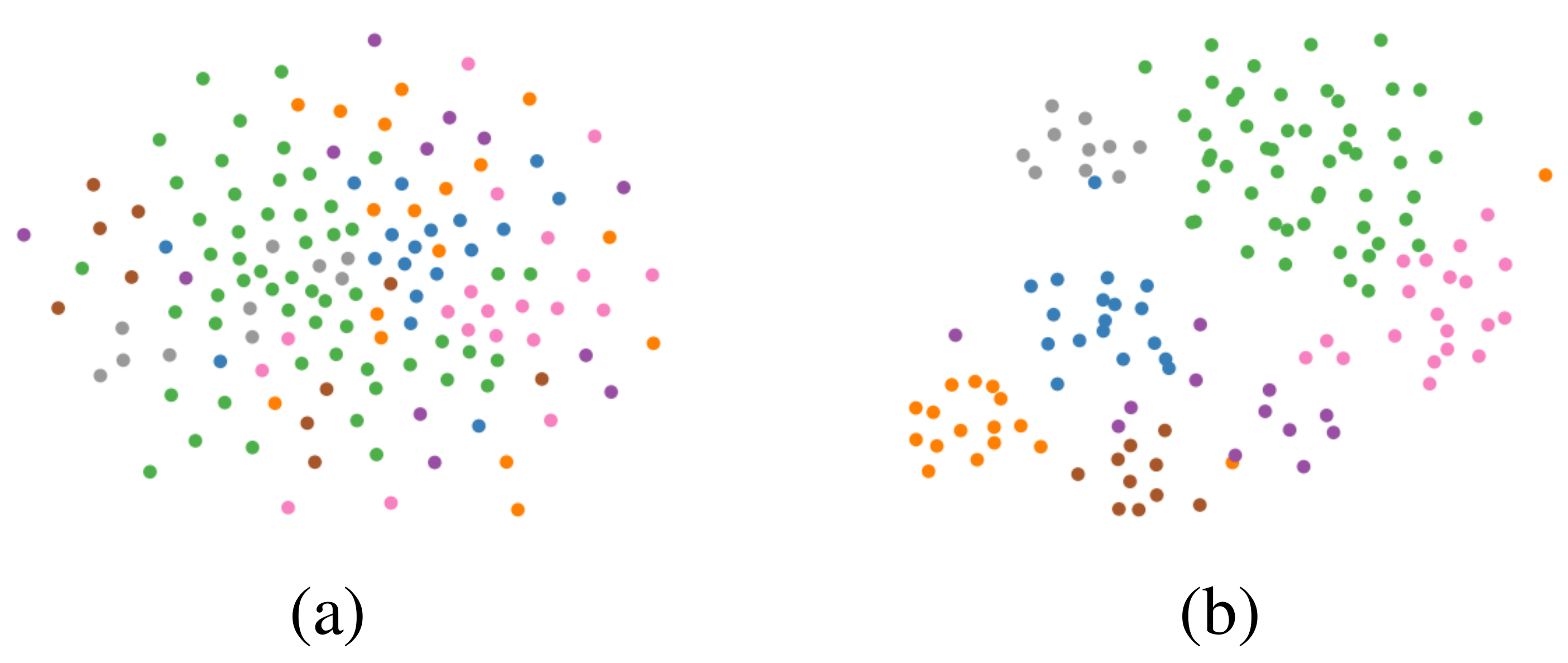}
\caption{T-SNE visualization of the label representations on WOS dataset. Dots with same color are labels with a same father. (a) BERT model. (b) Our approach.}
\label{fig:5}
\end{figure}

Our approach attempts to incorporating hierarchy into the text representation, which is fed into a linear layer for probabilities as in Equation \ref{eq:3}. The weight matrix $W_C$ can be viewed as label representations and we plot theirs T-SNE projections under default configuration. Since a label and its father should be classified simultaneously, the representation of a label and its father should be similar. Thus, if the hierarchy is injected into the text representation, labels with the same father should have more similar representation to each other than those with a different father. As illustrated in Figure \ref{fig:5}, label representations of BERT are scattered while label representations of our approach are clustered, which demonstrates that our text encoder can learn a hierarchy-aware representation.

\begin{table}[t]
\centering
\resizebox{0.9\linewidth}{!}{
\begin{tabular}{lcc}
\toprule
Variants of Graphormer     & Micro-F1 & Macro-F1 \\ \midrule
Base architecture     & \textbf{87.46}    & \textbf{81.52}  \\
-w/o name embedding   & 86.40    &  80.40  \\
-w/o spatial encoding & 86.88   &  80.42   \\
-w/o edge encoding    & 87.25    & 80.54    \\ \bottomrule
\end{tabular}}
\caption{Performance with variants of Graphormer on development set of WOS. We remove name embedding, spatial encoding, and edge encoding respectively. ``w/o'' stands for ``without''.}
\label{tab:4}
\end{table}

\begin{table}[t]
\centering
\resizebox{0.9\linewidth}{!}{
\begin{tabular}{lcc}
\toprule
Generation Strategy     & Micro-F1 & Macro-F1 \\ \midrule
Hierarchy-guided     & \textbf{87.46}    & \textbf{81.52}  \\
Dropout   & 86.94  & 79.91\\
Random masking &  87.19   &  81.16  \\ 
Adversarial attack & 86.67 & 80.24 \\ \bottomrule
\end{tabular}}
\caption{Impact of different positive example generation techniques on the development set of WOS. Hierarchy-guided is the proposed method. We control the valid tokens in positive samples roughly the same for random methods. We select FGSM as the attack algorithm following \citet{pan2021improved}.}
\label{tab:5}
\end{table}

\subsubsection{Effect of Graphormer}

As for the components of the Graphormer, we validate the utility of name embedding, spatial encoding, and edge encoding. As in Table \ref{tab:4}, all three components contribute to embedding the graph. Edge encoding is the least useful among these three components. Edge encoding is supposed to model the edge features provided by the graph, but the hierarchy of HTC has no such information so that the effect of edge encoding is not fully embodied in this task. Name embedding contributes most among components. Previous work only initialize embedding weights with label name but we treat it as a part of input features. As a result, neglecting name embedding observes the largest drop, which may explain the poor performance on RCV1-V2.

\subsubsection{Effect of Positive Example Generation}
To further illustrate the effect of our data generation approach, we compare it with a few generation strategies. Dropout \cite{gao2021simcse} uses no positive sample generation techniques but contrasts on the randomness of the Dropout function using two identical models. Random masking \cite{meng2021coco} is similar to our approach except the remained tokens are randomly selected. Adversarial attack \cite{pan2021improved} generates positive examples by an attack on gradients.

As in Table \ref{tab:5}, a duplication of the model as positive examples is effective but performs poorly. Instead of dropping information at neuron level, random masking drops entire tokens and boosts Macro-F1 by over 1\%, indicating the necessity of building hard enough contrastive examples. The adversarial attack can build hard-enough samples by gradient ascending and disturbance in the embedding space. But the disturbance is not regularized by hierarchy or labels so that it is less effective since there is no guarantee that the adversarial examples remain the label. Our approach guided the example construction by both the hierarchy and the labels, which accommodates with HTC most and achieves the best performance.

\begin{figure}[t]
\centering
\includegraphics[width=\linewidth]{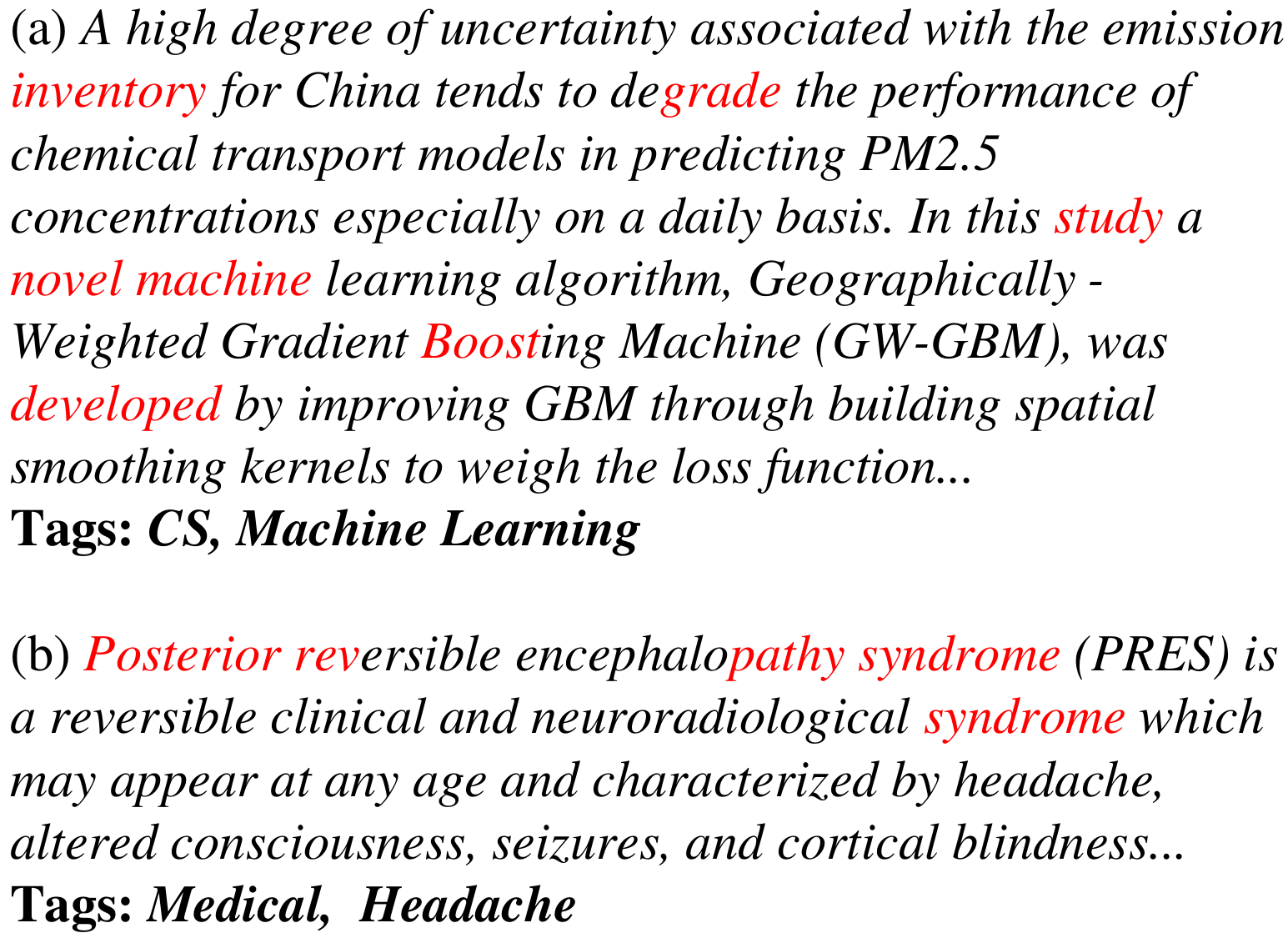}
\caption{Two fragments of the generated positive examples. Tokens in red are kept for positive examples. We omit a few unrelated tokens (such as \textit{a}, \textit{the}, or comma) for clarity.}
\label{fig:3}
\end{figure}

In Figure \ref{fig:3}, we select two cases to further illustrate the effect of labels on positive samples generation. In the first case, word \textit{machine} strongly indicates this passage belongs to \textit{Machine Learning} so that it is kept for positive examples. In the second case, \textit{syndrome} is related to \textit{Medical} and \textit{PRES} occurs several times among \textit{Headache}. Because of the randomness of sampling, our approach cannot construct an example with all keywords. For instance, \textit{learning} in case one or \textit{headache} in case two is omitted in this trial, which adds more difficulties for contrastive examples.


\section{Conclusion}
In this paper, we present Hierarchy-guided Contrastive Learning (HGCLR) for hierarchy text classification. We adopt contrastive learning for migrating taxonomy hierarchy information into BERT encoding. To this end, we construct positive examples for contrastive learning under the guidance of a graph encoder, which learns label features from taxonomy hierarchy. We modify Graphormer, a state-of-the-art graph encoder, for better graph understanding. Comparing to previous approaches, our approach empirically achieves consistent improvements on two distinct datasets and comparable results on another one. All of the components we designed are proven to be effective.

\section*{Acknowledgements}
We thank all the anonymous reviewers for their constructive feedback. The work is supported by National Natural Science Foundation of China under Grant No.62036001 and PKU-Baidu Fund (No. 2020BD021).

\bibliography{acl_latex}
\bibliographystyle{acl_natbib}

\appendix

\section{Trick for Token Selection} \label{app:1}
To make sure $P_i$ in Equation \ref{eq:2} can acquire gradients, we choose to modify token embedding instead of the token itself. As in Equation \ref{eq:6}, $e_i$ is the token embedding of $x_i$ and can have gradient. The positive counterpart of $e_i$ is denoted as:
\begin{equation}
    \hat{e_i}=e_i((P_i+Detach(1-P_i)) \text{ if } P_i>\gamma \text{ else } 0 ),
\end{equation}
where $Detach$ is a function that ignores the gradient of its input. Numerically, $\hat{e_i}$ is either $e_i$ or $0$ depending on the threshold $\gamma$, which serves the same purpose as Equation \ref{eq:2}. As for gradient, 
\begin{equation}
    \frac{\partial \hat{e_i}}{\partial P_i}=e_i \text{ if } P_i>\gamma \text{ else } 0,
\end{equation}
which makes $P_i$ can be updated by back-propagation.



\end{document}